\documentclass[letterpaper]{article} 
\usepackage{aaai2027}  
\usepackage{times}    
\usepackage{helvet}   
\usepackage{courier}  

\usepackage[hyphens]{url} 
\usepackage{graphicx}     
\usepackage{natbib}  
\usepackage{caption} 
\usepackage{subcaption}

\usepackage{amsmath}
\usepackage{amssymb}
\usepackage{amsfonts}

\usepackage{tabularx}
\usepackage{array}
\usepackage{booktabs}
\usepackage{multirow}

\usepackage{algorithm}
\usepackage{algorithmic}

\usepackage{newfloat}
\usepackage{listings}
\DeclareCaptionStyle{ruled}{labelfont=normalfont,labelsep=colon,strut=off} 
\floatstyle{ruled}
\newfloat{listing}{tb}{lst}{}
\floatname{listing}{Listing}

\usepackage{booktabs}

\title{\textsc{PARALLEL}: A Prefrontal-Aligned Reinforcement inspired Approach for Language-Model Learning under Explicit Limits}

\author{
    Namkyung Yoon\textsuperscript{\rm 1}\equalcontrib,
    Sanghong Kim\textsuperscript{\rm 1}\equalcontrib,
    Hwangnam Kim\textsuperscript{\rm 1}\equalcontrib\corresponding
}

\affiliations{
    \textsuperscript{\rm 1}School of Electrical Engineering, Korea University\\
    Seoul 02841, Republic of Korea\\
    nkyoon93@korea.ac.kr,
    sanghongkim@korea.ac.kr,
    hnkim@korea.ac.kr
}

\begin{document}

\maketitle

\begin{abstract}
Recent language models achieve strong performance across a variety of tasks, but conventional adaptation applies updates uniformly across training samples regardless of their local update benefit. 
We propose \textsc{PARALLEL}, a prefrontal-aligned reinforcement inspired approach for language-model learning. Inspired by the complementary roles of goal-related and uncertainty-related control, PARALLEL represents these forms of information as separate controller signals and combines them with the current model representation. 
A reinforcement-inspired controller assigns sample-dependent update intensity using immediate utility--cost feedback.
\textsc{PARALLEL} therefore learns when and how strongly to adapt to each sample, prioritizing beneficial updates while limiting unnecessary parameter changes. \textsc{PARALLEL} uses available updates more efficiently than selective baselines while retaining $94.1$--$99.2\%$ of Full-adaptation performance. Beyond multiple-choice reasoning, experiments on XSum and CNN/DailyMail show that \textsc{PARALLEL} retains $96.9$--$98.6\%$ of the ROUGE-1 and ROUGE-2 scores achieved by Full adaptation and $98.8$--$98.9\%$ of the corresponding ROUGE-L scores. When compared at the same cumulative adaptation time or GPU energy, \textsc{PARALLEL} achieves higher ARC accuracy and exhibits a more stable late-stage adaptation trajectory than Full adaptation in the representative run. These results show that learning when and how strongly to update each sample supports stable and efficient post-deployment stream adaptation while avoiding unnecessary updates.
\end{abstract}


\section{Introduction}

Recent language models achieve strong performance across a wide range of tasks, including natural language understanding, reasoning, and knowledge-intensive question answering ~\cite{zhao2026survey,chang2024survey}. However, adapting these models to task-specific data still requires repeated parameter updates. Conventional learning methods generally apply updates uniformly across training samples, even though the local benefit of an update can vary across samples. Consequently, repeatedly updating low-utility samples may consume adaptation time and energy without proportional performance improvement~\cite{yoon2026beyond}.

This limitation is particularly relevant to post-deployment stream adaptation: a general-purpose language model may require supervised adaptation to a target-task distribution, yet applying full-intensity updates to every incoming labeled sample can accumulate substantial time and energy without proportional performance gains~\cite{hoi2021online}. Our objective is therefore to allocate adapter updates across the stream according to their estimated utility, supporting effective target-task adaptation while avoiding unnecessary updates. 

Existing approaches improve adaptation efficiency through different strategies. Parameter-efficient methods reduce the number of trainable parameters~\cite{hu2022lora}. Data-selection methods instead prioritize samples based on uncertainty or difficulty~\cite{settles2009active,paul2021deep}. However, parameter-efficient adaptation does not control how updates are allocated across samples, while data-selection methods typically apply the same update strength to every selected sample. Consequently, neither approach jointly determines whether an update is needed and how strongly each incoming sample should influence adaptation.
To address this problem, we draw inspiration from an organizational principle reported in the human lateral prefrontal cortex. Recent neuroimaging evidence indicates that goal-related and uncertainty-related information is represented in a factorized manner, supporting stable yet flexible learning~\cite{sung2025factorized}. This finding motivates a control structure in which goal-related and uncertainty-related signals are modeled separately and then combined to regulate language-model adaptation. 
\begin{figure*}[t]
    \centering
    \includegraphics[width=\textwidth]{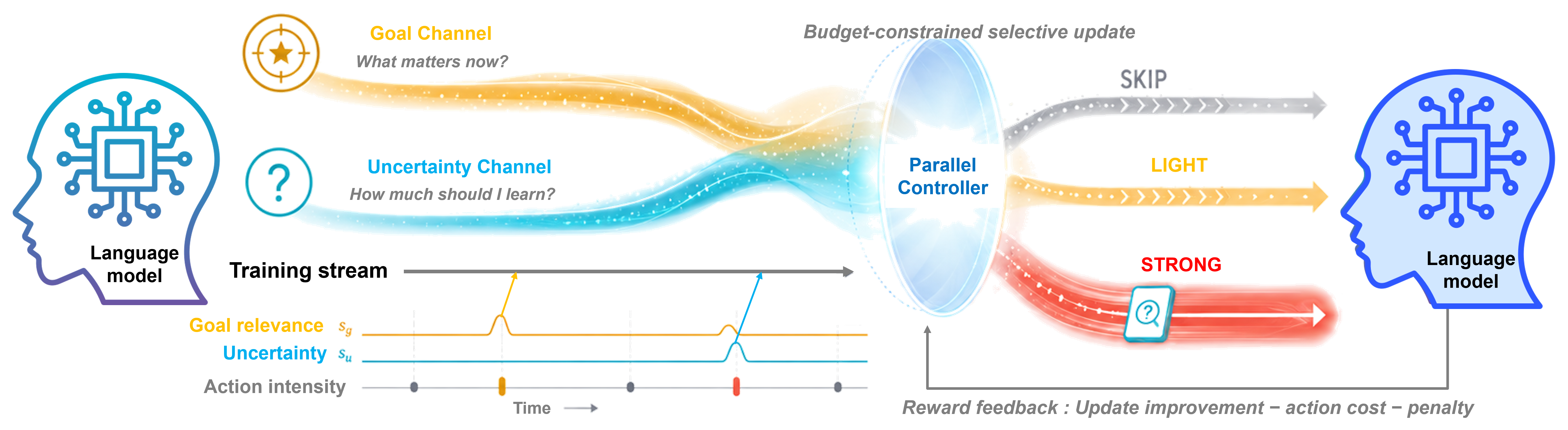}
    \caption{Overview of the proposed PARALLEL framework.}
    \label{fig:parallel_overview}
\end{figure*}






Based on this principle, we propose \textsc{PARALLEL}, a Prefrontal-Aligned Reinforcement Approach for Language-Model Learning under Explicit Limits. Rather than applying uniform updates throughout the stream, \textsc{PARALLEL} learns whether and how strongly the model should adapt to each incoming sample, thereby supporting efficient post-deployment adaptation while avoiding unnecessary updates. PARALLEL represents goal-related and uncertainty-related information as separate controller signals and combines them with the current model representation. 
A controller trained with one-step REINFORCE uses these signals to determine whether and how strongly to update the adapter on each labeled sample. It learns from immediate utility--cost feedback, while a hard cumulative update-mass constraint bounds the total update intensity allocated across the stream.

We evaluate \textsc{PARALLEL} on three reasoning and two summarization benchmarks. The main experiments compare \textsc{PARALLEL} with Frozen and Full adaptation. Additional budget-matched comparisons are conducted across all five benchmarks, while cross-backbone evaluation, controller-input ablations, and stream-level measurements of adaptation time and GPU energy are conducted in the benchmark setting. The results show that \textsc{PARALLEL} effectively allocates a bounded update budget and retains most of the performance under Full adaptation across three compact backbones. Accordingly, \textsc{PARALLEL} should be viewed as a mechanism for controlling update allocation rather than accelerating offline fine-tuning.

The main contributions are as follows:
\begin{itemize}
    \item We propose \textsc{PARALLEL}, which uses separate goal-related and uncertainty-related signals to perform online LoRA update control through three actions, determining whether and how strongly to update each incoming sample under a cumulative update-mass constraint.
    \item We demonstrate effective budget-constrained adaptation across three reasoning and two summarization benchmarks using multiple language models.
\end{itemize}

The remainder of this paper is organized as follows. Section~2 reviews related work, Section~3 presents PARALLEL, Section~4 reports the experimental evaluation, and Section~5 concludes the paper.

\section{Related Work}

\subsection{Efficient Adaptation Methods}

Full-parameter fine-tuning updates all parameters of a pretrained language model and therefore requires substantial memory and computation~\cite{devlin2019bert}. Parameter-efficient methods instead reduce the number of trainable parameters while preserving the pretrained backbone. LoRA introduces trainable low-rank components while keeping the original model parameters frozen~\cite{hu2022lora}. In our evaluation, \emph{Full adaptation} applies the same LoRA configuration to every sample in the adaptation stream.

Data-selection strategies determine which samples are used for adaptation under a limited sample budget. \emph{Random selection} draws a fixed-size subset uniformly at random~\cite{citovsky2021batch}. Uncertainty-based active learning prioritizes samples on which the current model is least confident~\cite{lewis1994sequential,settles2009active}. Its entropy-based variant ranks samples by predictive entropy and selects those with the highest values~\cite{settles2009active}. 
Parameter-efficient methods control which parameters are trained, whereas data-selection methods control which samples are selected. PARALLEL instead processes the labeled adaptation stream sequentially and learns sample-wise update intensity, while a cumulative constraint prevents excessive updates.

\subsection{Reinforcement Learning for Adaptation Control}

Reinforcement learning has been widely applied to language models to align generated responses with task objectives, human preferences, and instruction-following requirements~\cite{ouyang2022training}. In conventional reinforcement learning from human feedback, the language model defines a policy over generated tokens or responses, and reward feedback directly optimizes its generation behavior.

PARALLEL instead adopts a reinforcement-inspired controller for adaptation control. For each labeled sample, the controller selects the intensity of the supervised adapter update and learns from immediate utility--cost feedback. The adapter itself is optimized only with the supervised task loss.


\subsection{Prefrontal-Aligned Learning Control}

Recent neuroimaging evidence suggests that goal-related and uncertainty-related information is represented in a factorized manner within the lateral prefrontal cortex, supporting stable yet flexible learning~\cite{sung2025factorized}. This organization motivates treating information about the current task objective and predictive uncertainty as distinct but complementary signals for learning control.

PARALLEL adopts this principle as a computational abstraction. It represents goal-related and uncertainty-related information as separate controller signals and combines them with the current model representation to allocate update intensity across samples.

\section{System Design}
\label{sec:system_design}

As shown in Figure~\ref{fig:parallel_overview}, \textsc{PARALLEL} controls whether and how strongly the adapter is updated for each incoming labeled sample. For each sample, it computes a goal-related signal representing consistency with the target response and an uncertainty-related signal reflecting the reliability of the current prediction.

Our PARALLEL controller combines these signals with the current model representation and selects the update intensity. The pretrained language-model backbone remains frozen, while the controller uses the goal-related and uncertainty-related signals to determine whether and how strongly to update the LoRA adapter for each sample.

\subsection{Control Signals in \textsc{PARALLEL}}
\label{subsec:control_signals}

Let \(\mathcal{D}=\{(x_t,y_t^{\star})\}_{t=1}^{T}\) denote the labeled adaptation stream, where \(T\) is the total number of adaptation samples, \(x_t\) is the question or prompt of the \(t\)-th sample, and \(y_t^{\star}\) is its target response. The language-model backbone with parameters \(\theta\) remains frozen, while task adaptation is performed through the model with parameters \(\theta_a\). PARALLEL observes the labeled samples sequentially and determines whether and how strongly \(\theta_a\) should be updated for each sample.

Before choosing an update action, \textsc{PARALLEL} assesses how well the current model has already learned the labeled sample $(x_t, y_t^\star)$. It performs a diagnostic forward pass using the current model parameters without applying an update. The resulting prediction quality serves as goal-related evidence for the controller: poor prediction supports a stronger update, whereas accurate prediction supports a light update or skipping the sample.

Since the target response $y_t^\star$ is available during supervised adaptation, \textsc{PARALLEL} evaluates it using the standard autoregressive objective adopted in supervised fine-tuning of language models~\citep{chatterjee2025effect}. Specifically, each target token is evaluated using the input and the preceding ground-truth target tokens. The diagnostic forward pass produces the following length-normalized pre-update target loss:
\begin{equation}
\ell_t^{-}
=
-\frac{1}{N_t}
\sum_{j=1}^{N_t}
\log p_{\theta,\theta_a}
\left(
y_{t,j}^{\star}
\mid
x_t, y_{t,<j}^{\star}
\right),
\end{equation}
where $N_t$ is the number of target tokens and the superscript $-$ denotes measurement before the update action is selected and applied. Length normalization makes the loss comparable across responses of different lengths. A small $\ell_t^{-}$ means that the model already assigns high probability to the target response, while a large $\ell_t^{-}$ indicates a greater need for adaptation. This loss is used only to guide action selection during supervised adaptation and is not required during inference.

As a complementary discrete measure, the same diagnostic forward pass also computes the pre-update target-token accuracy:
\begin{equation}
A_t^{-}=\frac{1}{N_t}\sum_{j=1}^{N_t}\mathbb{I}\left[\arg\max_{y}p_{\theta,\theta_a}\left(y\mid x_t,y_{t,<j}^{\star}\right)=y_{t,j}^{\star}\right],
\label{eq:pre_update_accuracy}
\end{equation}
where \(\mathbb{I}[\cdot]\) is the indicator function. Thus, \(A_t^{-}\) is the proportion of target positions at which the most probable token matches the ground-truth token. The loss is a continuous measure that reflects the probability assigned to each target token, whereas target-token accuracy is an indicator-based measure of whether the top-1 prediction is correct.

PARALLEL constructs the goal-related signal as
\begin{equation}
s_{g,t}
=
\left[
A_t^{-},
\exp(-\ell_t^{-})
\right],
\label{eq:goal_signal}
\end{equation}
where $A_t^{-}$ denotes the target-token accuracy and $\ell_t^{-}$ is the length-normalized target loss before adaptation. Because $\ell_t^{-}$ is length-normalized, $\exp(-\ell_t^{-})$ corresponds to the geometric mean of the probabilities assigned to the target tokens and therefore serves as a probability-based target-confidence measure. Together, these two components summarize how closely the current model behavior agrees with the labeled target.

Target agreement alone does not describe how confidently the model forms its prediction. PARALLEL therefore constructs an uncertainty-related signal containing five complementary statistics:
\begin{equation}
s_{u,t}=\left[H_t^{\mathrm{choice}},H_t^{\mathrm{tok}},c_t,m_t,v_t\right].
\label{eq:uncertainty_signal}
\end{equation}
The five components are defined as follows:
\begin{itemize}
    \item \(H_t^{\mathrm{choice}}\) is the normalized entropy over candidate responses. A high value indicates that probability is distributed across multiple choices.
    \item \(H_t^{\mathrm{tok}}\) is the average normalized entropy of the token-level predictive distributions. A high value indicates diffuse token predictions.
    \item \(c_t\) is the mean top-1 confidence across target-token positions. A high value indicates strong confidence in the most probable tokens.
    \item \(m_t\) is the mean probability margin between the top-1 and top-2 tokens. A large margin indicates a more decisive prediction.
    \item \(v_t\) is the variance of the probabilities assigned to the target tokens across positions. It represents how consistently the model assigns confidence throughout the target response.
\end{itemize}
The normalized token entropy \(H_t^{\mathrm{tok}}\) is computed by dividing the predictive entropy at each target-token position by \(\log|\mathcal{V}|\), where \(\mathcal{V}\) is the model vocabulary, and then averaging over the target positions.

For a multiple-choice sample with \(K_t\) candidate responses, let \(\ell_{t,k}^{\mathrm{cand}}\) denote the length-normalized loss of candidate \(k\). Its choice probability is defined as follows:
\begin{equation}
\begin{aligned}
q_{t,k}
&=
\frac{\exp(-\ell_{t,k}^{\mathrm{cand}})}
{\sum_{r=1}^{K_t}\exp(-\ell_{t,r}^{\mathrm{cand}})},\\
H_t^{\mathrm{choice}}
&=
-\frac{1}{\log K_t}
\sum_{k=1}^{K_t}q_{t,k}\log q_{t,k}.
\end{aligned}
\label{eq:choice_entropy}
\end{equation}
For XSum and CNN/DailyMail, which do not provide discrete answer choices, we set \(H_t^{\mathrm{choice}}=0\), and the controller relies on the remaining goal-related, uncertainty-related, and representation signals.
Lower-loss candidates receive higher probabilities, while
$H_t^{\mathrm{choice}}$ measures the ambiguity among the candidates.
The normalization by \(\log K_t\) makes the entropy comparable across samples with different numbers of choices. \(H_t^{\mathrm{choice}}\) is high when probability is distributed across several candidates and low when the prediction is concentrated on a particular candidate. The goal-related signal therefore describes how closely the prediction agrees with the target, whereas the uncertainty-related signal describes how confidently the prediction is formed.

The two signals contain numerical summaries of model behavior but do not fully represent the meaning or context of the sample. To provide this information, PARALLEL concatenates them with the final prompt-token representation \(h_t\):
\begin{equation}
z_t=[s_{g,t};s_{u,t};h_t],
\label{eq:controller_state}
\end{equation}
where the semicolon denotes concatenation. The resulting state combines target agreement, predictive uncertainty, and the current model representation. All components of \(z_t\) are detached from the language-model computation graph before being passed to the controller.

\subsection{Budget-Constrained Controller Learning}
\label{subsec:budgeted_controller}

PARALLEL uses the controller state \(z_t\) to produce an action policy as follows:
\begin{equation}
\pi_{\theta_c}(a_t\mid z_t)
=
\left[\operatorname{softmax}\left(f_{\theta_c}(z_t)\right)\right]_{a_t}.
\label{eq:controller_policy}
\end{equation}
where \(f_{\theta_c}\) is a lightweight controller network with parameters \(\theta_c\), and \(a_t\) denotes the selected action for sample \(t\).

The goal-related and uncertainty-related signals are not independently mapped to separate actions. Instead, the controller interprets their joint evidence to determine the required degree of adaptation. When the current prediction is sufficiently aligned with the target and is considered reliable, the update can be inhibited. When the evidence is intermediate or the two signals provide conflicting information, a moderated adjustment can be applied. When the current behavior indicates a clear need for correction, the adapter can receive a full-strength update. From a cognitive-control perspective, these cases correspond to inhibition, moderated adjustment, and full correction.
PARALLEL represents these three functional regimes as \(\mathcal{A}=\{\textsc{Skip},\textsc{Light},\textsc{Strong}\}\). The update weight associated with the selected action is
\begin{equation}
w_t=w(a_t)=
\begin{cases}
0, & a_t=\textsc{Skip},\\
\lambda_t, & a_t=\textsc{Light},\\
1, & a_t=\textsc{Strong},
\end{cases}
\qquad 0<\lambda_t<1.
\label{eq:action_weight}
\end{equation}
Where \textsc{Skip} inhibits the adapter update, \textsc{Light} applies an intermediate update scale, and \textsc{Strong} applies the reference full-strength update. The controller learns how the combined state \(z_t\) maps to these regimes without relying on manually defined thresholds.

$r_t$ denotes the controller reward, which is formally defined in Eq.~\ref{eq:controller_reward}.
The \textsc{Light} scale \(\lambda_t\) is adjusted using the observed rewards of the \textsc{Light} and \textsc{Strong} actions. For \(k\in\{\textsc{Light},\textsc{Strong}\}\), the action-wise reward estimate is updated as
\begin{equation}
\bar{r}_{k,t}=
\begin{cases}
\beta_{\lambda}\bar{r}_{k,t-1}+(1-\beta_{\lambda})r_t, & a_t=k,\\
\bar{r}_{k,t-1}, & a_t\neq k.
\end{cases}
\label{eq:action_reward_ema}
\end{equation}
When both reward estimates are available, PARALLEL computes a target scale by comparing their positive reward per unit update mass:
\begin{equation}
\widehat{\lambda}_t
=
\operatorname{clip}\left(
\frac{[\bar{r}_{\mathrm{Light},t}]_{+}}
{\lambda_t\max([\bar{r}_{\mathrm{Strong},t}]_{+},\epsilon)},
\lambda_{\min},
\lambda_{\max}
\right).
\label{eq:target_light_scale}
\end{equation}

Because $\lambda_t$ is bounded below by $\lambda_{\min}>\epsilon$, the stabilizer is required only for the \textsc{Strong} return.
When the estimated \textsc{Strong} return is non-positive while the \textsc{Light} return is positive, the small stabilized denominator drives $\widehat{\lambda}_t$ toward $\lambda_{\max}$.
When both returns are non-positive, the ratio becomes zero and $\widehat{\lambda}_t$ is set to $\lambda_{\min}$.

The \textsc{Light} scale is then updated as
\begin{equation}
\lambda_{t+1}=
\operatorname{clip}\left(
\beta_{\lambda}\lambda_t+(1-\beta_{\lambda})\widehat{\lambda}_t,
\lambda_{\min},\lambda_{\max}
\right).
\label{eq:adaptive_light}
\end{equation}
The update of \(\lambda_t\) begins only after both \textsc{Light} and \textsc{Strong} have been selected at least once; until then, \(\lambda_t\) remains unchanged. The reported experiments use \(\lambda_1=0.5\), \(\beta_{\lambda}=0.95\), and \([\lambda_{\min},\lambda_{\max}]=[0.05,0.95]\).

The cumulative amount of allocated updating after processing \(t\) samples is
\begin{equation}
M_t=\sum_{i=1}^{t}w(a_i).
\label{eq:cumulative_update_mass}
\end{equation}

The cumulative update mass and its budget constraint are defined as
\begin{equation}
M_t=\sum_{i=1}^{t}w_i,\qquad \frac{M_T}{T}\leq\rho.
\label{eq:update_budget}
\end{equation}
Update mass is the cumulative sum of the action weights $w(a_t)$, and a budget ratio $\rho$ permits at most $\rho T$ mass over a $T$-sample stream. Before sampling, actions that would exceed the remaining mass are masked and the remaining probabilities are renormalized, ensuring $M_T \leq \rho T$. This budget limits the cumulative update intensity rather than the number of labeled samples processed, the number of LoRA parameters, or the measured parameter displacement.
When $w_t>0$, PARALLEL performs one AdamW step on the supervised target loss with the learning rate scaled by $w_t$. When $w_t=0$, the optimizer step is omitted. The language-model backbone remains frozen, and only the LoRA adapter is updated.
After applying the selected action, PARALLEL reevaluates the same sample to obtain the post-action target loss $\ell_t^{+}$ and target-token accuracy $A_t^{+}$. The immediate improvement is defined as
\begin{equation}
\Delta_t=(\ell_t^{-}-\ell_t^{+})+\gamma_{\mathrm{acc}}(A_t^{+}-A_t^{-}),
\label{eq:local_improvement}
\end{equation}
where $\gamma_{\mathrm{acc}}$ controls the contribution of the accuracy change. For \textsc{Skip}, no adapter update occurs, so $\ell_t^{+}=\ell_t^{-}$, $A_t^{+}=A_t^{-}$, and $\Delta_t=0$.

PARALLEL also considers how quickly the available budget is consumed. The budget-pressure term is
\begin{equation}
P_t=\left[\frac{M_t}{\rho T}-\frac{t}{T}\right]_{+}.
\label{eq:budget_pressure}
\end{equation}
The first fraction represents the consumed portion of the update budget, while the second represents the processed portion of the adaptation stream. Thus, \(P_t\) becomes positive when budget consumption advances faster than stream progression.

The controller reward is
\begin{equation}
r_t=\Delta_t-\alpha_c w_t-\alpha_b P_t,
\label{eq:controller_reward}
\end{equation}
where \(\alpha_c\) penalizes update intensity and \(\alpha_b\) penalizes excessive early budget consumption. The hard constraint in Eq.~\eqref{eq:update_budget} prevents budget violation, whereas \(P_t\) encourages the controller to distribute the available budget across the stream.

The controller is optimized using one-step REINFORCE~\cite{williams1992simple}:
\begin{equation}
\mathcal{L}_{\mathrm{ctrl}}
=-(r_t-b_{t-1})\log\widetilde{\pi}_{\theta_c,t}(a_t\mid z_t)
-\alpha_{\mathrm{ent}}\mathcal{H}\left(\widetilde{\pi}_{\theta_c,t}\right),
\label{eq:controller_objective}
\end{equation}
where \(\widetilde{\pi}_{\theta_c,t}\) is the budget-masked policy and \(\alpha_{\mathrm{ent}}\) controls entropy regularization. After each controller step, the reward baseline is updated using an exponential moving average, $b_t=\beta_b b_{t-1}+(1-\beta_b)r_t$. If $r_t>b_{t-1}$, the selected action is reinforced under similar controller states; otherwise, its probability is reduced. Entropy regularization prevents premature concentration on a single action.
Because \(r_t\) compares the same sample before and after the selected action, the controller learns from immediate utility--cost feedback rather than optimizing a long-horizon episodic return.

\begin{algorithm}[t]
\caption{\textsc{PARALLEL} adaptation}
\label{alg:parallel}
\begin{algorithmic}[1]
\REQUIRE Labeled adaptation stream $\mathcal{D}=\{(x_t,y_t^\star)\}_{t=1}^{T}$, frozen backbone $\theta$, LoRA adapter $\theta_a$, controller $\theta_c$, budget ratio $\rho$
\STATE Initialize $M_0\leftarrow0$, reward baseline $b_0$, Light scale $\lambda_1$, and action-reward estimates
\FOR{$t=1,\ldots,T$}
    \STATE Compute $\ell_t^{-}$, $A_t^{-}$, $s_{g,t}$, $s_{u,t}$, and $h_t$
    \STATE Form $z_t=[s_{g,t};s_{u,t};h_t]$ and compute $\pi_{\theta_c}(\cdot\mid z_t)$
    \STATE Mask actions satisfying $M_{t-1}+w_{\lambda_t}(a)>\rho T$ and renormalize to $\widetilde{\pi}_{\theta_c,t}$
    \STATE Sample $a_t\sim\widetilde{\pi}_{\theta_c,t}$ and set $w_t\leftarrow w_{\lambda_t}(a_t)$
    \IF{$w_t>0$}
        \STATE Update $\theta_a$ by one supervised AdamW step with the learning rate scaled by $w_t$
    \ENDIF
    \STATE Set $M_t\leftarrow M_{t-1}+w_t$
    \STATE Obtain $\ell_t^{+}$ and $A_t^{+}$, and compute reward $r_t$ using $M_t$
    \STATE Update $\theta_c$ using Eq.~\eqref{eq:controller_objective} with baseline $b_{t-1}$
    \STATE Update the action-reward estimates and Light scale $\lambda_{t+1}$
    \STATE Update $b_t$
\ENDFOR
\RETURN Adapted LoRA parameters $\theta_a$
\end{algorithmic}
\end{algorithm}


\begin{table}[t]
\centering
{\footnotesize
\setlength{\tabcolsep}{3pt}
\renewcommand{\arraystretch}{1.03}
\begin{tabular}{@{}llccc@{}}
\toprule
Dataset & Metric & Frozen & \textsc{PARALLEL} & Full LoRA \\
\midrule
ARC-Challenge & \texttt{acc\_norm} & 0.481 & 0.569 & 0.584 \\
OpenBookQA & \texttt{acc\_norm} & 0.418 & 0.540 & 0.574 \\
CommonsenseQA & Accuracy & 0.789 & 0.842 & 0.849 \\
\midrule
XSum & R-1 & 0.2000 & 0.3650 & 0.3723 \\
     & R-2 & 0.0506 & 0.1409 & 0.1454 \\
     & R-L & 0.1392 & 0.2915 & 0.2951 \\
\midrule
CNN/DM & R-1 & 0.3067 & 0.3937 & 0.3991 \\
       & R-2 & 0.1034 & 0.1694 & 0.1742 \\
       & R-L & 0.1870 & 0.2685 & 0.2715 \\
\bottomrule
\end{tabular}
}
\caption{Overall benchmark results.}
\label{tab:overall_task_performance}
\end{table}

\section{Experiments}
\label{sec:experiments}

\subsection{Experimental Setup}

We evaluate \textsc{PARALLEL} on ARC-Challenge~\cite{clark2018think}, OpenBookQA~\cite{mihaylov2018can}, CommonsenseQA~\cite{talmor2019commonsenseqa}, XSum~\cite{narayan2018xsum}, and CNN/DailyMail~\cite{hermann2015cnn}. The first three benchmarks evaluate science and commonsense reasoning, while XSum and CNN/DailyMail evaluate abstractive news summarization in separate experiments.

The primary evaluation uses Qwen2.5-3B-Instruct on all five benchmarks. For the reasoning experiments, fixed training streams contain 90\% of each official training split: 1,007 examples for ARC-Challenge, 4,461 for OpenBookQA, and 8,767 for CommonsenseQA. 

We additionally evaluate Qwen2.5-0.5B-Instruct, SmolLM2-360M-Instruct~\cite{allal2025smollm2}, and TinyLlama-1.1B-Chat~\cite{zhang2024tinyllama} on ARC-Challenge with the official training split and seeds 42, 43, and 44.
All adaptation methods use a frozen language-model backbone and a trainable LoRA adapter~\cite{hu2022lora}. Training uses one pass, batch size 1, bfloat16, and AdamW with a learning rate of \(2\times10^{-5}\), linear scheduling, and a warmup ratio of 0.03. The LoRA configuration uses rank 16, scaling 32, and dropout 0.05. Within each model--dataset condition, all methods use the same tokenizer, prompt format, adapter configuration, and optimizer settings.

The PARALLEL controller is a one-hidden-layer multilayer perceptron with hidden dimension 64 and a GELU activation. It is optimized using AdamW with a learning rate of \(8\times10^{-4}\). The controller objective uses \(\gamma_{\mathrm{acc}}=0.30\), \(\alpha_c=0.14\), \(\alpha_b=0.40\), \(\alpha_{\mathrm{ent}}=0.01\), and reward-baseline decay \(\beta_b=0.95\). The LoRA adapter is optimized using the supervised target loss, whereas the controller is optimized using reward feedback. 

Final adapters are evaluated without in-context demonstrations. For the reasoning benchmarks, we use \texttt{lm-eval-harness}~\cite{biderman2024lessons} and report \texttt{acc\_norm} for ARC-Challenge and OpenBookQA and \texttt{acc} for CommonsenseQA. For XSum and CNN/DailyMail, we report ROUGE-1, ROUGE-2, and ROUGE-L.

\begin{table}[t]
\centering
{\footnotesize
\setlength{\tabcolsep}{5.0pt}
\renewcommand{\arraystretch}{1.05}
\begin{tabular}{@{}cccc@{}}
\toprule
\textbf{Target \(\rho\)} & \textbf{ARC-C} & \textbf{OBQA} & \textbf{CSQA} \\
\midrule
\(0.15\) & \(0.5162\) & \(0.4960\) & \(0.8378\) \\
\(0.30\) & \(0.5691\) & \(0.5400\) & \(0.8419\) \\
\(0.50\) & \(0.5538\) & \(0.5600\) & \(0.8403\) \\
\bottomrule
\end{tabular}
}
\caption{Budget sensitivity of \textsc{PARALLEL}.}
\label{tab:budget_sensitivity}
\end{table}

\subsection{Compared Methods and Metrics}

Frozen performs no adaptation, while Full adaptation applies the same LoRA update to every sample. Across all five benchmarks, we additionally compare two budget-matched subset baselines: \emph{Random selection}~\cite{citovsky2021batch}, entropy-based \emph{Active learning}~\cite{lewis1994sequential,settles2009active}. The selection scores are computed once before adaptation, and each selected sample receives a full-strength update. \textsc{PARALLEL} instead uses the goal-related and uncertainty-related signals to assign action-dependent update weights while processing the complete labeled stream.
Following Eq.~\eqref{eq:update_budget}, normalized update mass is reported as \(M_T/T\). Frozen and Full adaptation have masses of \(0\) and \(1\), respectively, and the main budget-matched comparison uses \(\rho=0.30\).

For a task score \(S\), we report
\begin{equation}
R_{\mathrm{score}}
=100 \times \frac{S}{S_{\mathrm{Full}}},
\qquad
R_{\mathrm{gain}}
=100 \times
\frac{S-S_{\mathrm{Frozen}}}
{S_{\mathrm{Full}}-S_{\mathrm{Frozen}}}.
\label{eq:relative_metrics}
\end{equation}
These metrics denote Full-score retention and adaptation-gain recovery, respectively.
For XSum and CNN/DailyMail, benchmark-level \(R_{\mathrm{gain}}\) is computed by averaging the values for ROUGE-1, ROUGE-2, and ROUGE-L.

\subsection{Budget Selection}
Before conducting the full evaluation, we perform a preliminary search to select an update budget for \textsc{PARALLEL}. Rather than tuning the budget separately for each dataset, we evaluate three representative ratios, \(\rho \in \{0.15, 0.30, 0.50\}\), on a subset of the benchmarks. The aim is to identify a practical operating point rather than a globally optimal ratio. As shown in Table~\ref{tab:budget_sensitivity}, increasing \(\rho\) from \(0.15\) to \(0.30\) improves performance across the evaluated benchmarks, whereas further gains at \(0.50\) are task dependent despite the additional updates. Among the evaluated ratios, \(\rho=0.30\) therefore provides the most favorable balance between performance and update cost.

\subsection{Experimental Results}
Table~\ref{tab:overall_task_performance} summarizes the results across reasoning and summarization benchmarks. On ARC-Challenge, OpenBookQA, and CommonsenseQA, \textsc{PARALLEL} retains $97.4\%$, $94.1\%$, and $99.2\%$ of the performance under Full adaptation, respectively. It also recovers $85.4\%$, $78.2\%$, and $88.3\%$ of the corresponding gains over Frozen. On XSum, performance retention ranges from $96.9\%$ to $98.8\%$, while gain recovery ranges from $95.3\%$ to $97.7\%$. On CNN/DailyMail, performance retention and gain recovery across the three ROUGE metrics range from 97.2\% to 98.9\% and from 93.2\% to 96.4\%, respectively.



\begin{table}[t]
\centering
{\footnotesize
\setlength{\tabcolsep}{3.2pt}
\renewcommand{\arraystretch}{1.02}
\begin{tabular}{@{}llccc@{}}
\toprule
\textbf{Task} & \textbf{Metric}
& \textbf{Random selection}
& \textbf{Active learning}
& \textbf{PARALLEL} \\
\midrule
ARC-C & Acc.
& $0.5452$ & $0.5213$ & $0.5691$ \\
OBQA & Acc.
& $0.5440$ & $0.5180$ & $0.5400$ \\
CSQA & Acc.
& $0.8313$ & $0.8444$ & $0.8419$ \\
\midrule
\multirow{3}{*}{XSum}
& R-1 & $0.3652$ & $0.3595$ & $0.3650$ \\
& R-2 & $0.1414$ & $0.1364$ & $0.1409$ \\
& R-L & $0.2914$ & $0.2863$ & $0.2915$ \\
\midrule
\multirow{3}{*}{CNN/DM}
& R-1 & $0.3950$ & $0.3926$ & $0.3937$ \\
& R-2 & $0.1705$ & $0.1647$ & $0.1694$ \\
& R-L & $0.2679$ & $0.2343$ & $0.2685$ \\
\bottomrule
\end{tabular}
}
\caption{Comparison of various adaptation strategies.}
\label{tab:primary_baselines}
\end{table}

\begin{figure}[t]
\centering
\includegraphics[width=1.0\columnwidth]{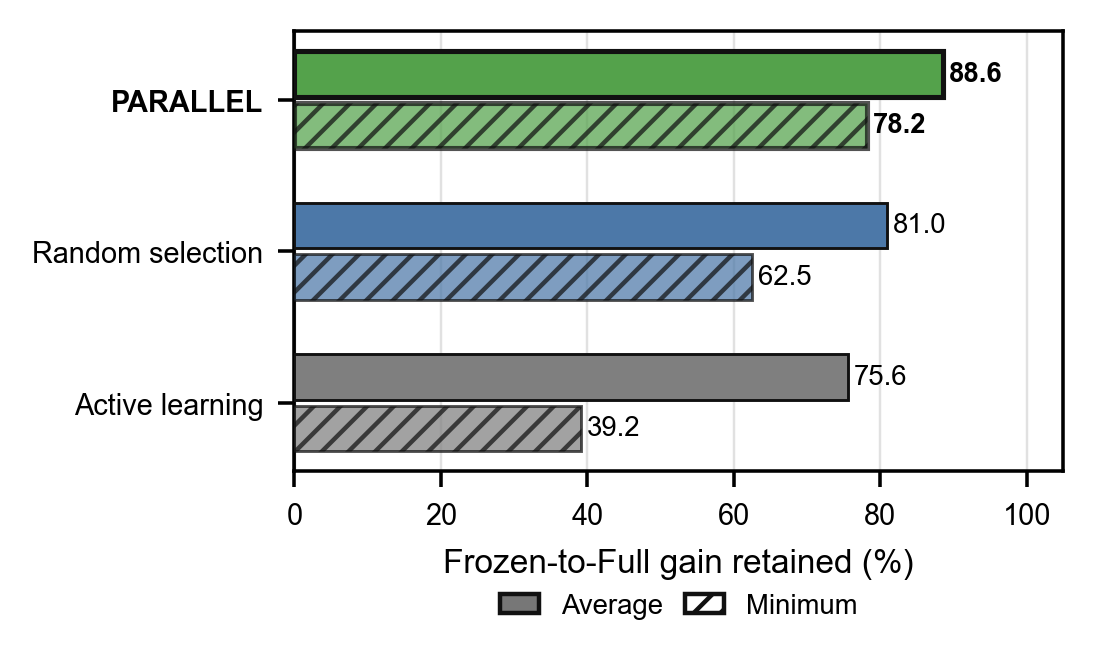}
\caption{Gain recovery of budget-matched methods across five benchmarks.}
\label{fig:gain_retention}
\end{figure}

Table~\ref{tab:primary_baselines} evaluates update-allocation efficiency by comparing \textsc{PARALLEL} with fixed subset-selection methods under the same target update-mass ratio.
Among the budget-matched methods, PARALLEL achieves the highest ARC-Challenge score, outperforming the next-best method by 0.0239. Overall, PARALLEL remains consistently at or near the top.
\textsc{PARALLEL} achieves the highest ROUGE-L on both XSum and CNN/DailyMail, with scores of $0.2915$ and $0.2685$. This consistent ROUGE-L advantage indicates better preservation of long-sequence content overlap under the same update budget. Consequently, Figure~\ref{fig:gain_retention} shows that \textsc{PARALLEL} achieves the highest average \(R_{\mathrm{gain}}\) recovery of 88.6\% across the five benchmarks, with a minimum benchmark-level recovery of 78.2\%.


\begin{table}[t]
\centering
{\footnotesize
\setlength{\tabcolsep}{2.0pt}
\renewcommand{\arraystretch}{1.08}
\begin{tabular*}{\columnwidth}{@{\extracolsep{\fill}}lcccc@{}}
\toprule
\textbf{Backbone} & \textbf{Frozen}
& \shortstack{\textbf{Full} \textbf{LoRA}}
& \textbf{PARALLEL} & \textbf{Mass} \\
\midrule
Qwen-0.5B & \(0.338\) & \(0.388{\pm}0.004\) & \(0.371{\pm}0.003\) & \(0.297\) \\
Smol-360M & \(0.344\) & \(0.401{\pm}0.001\) & \(0.386{\pm}0.004\) & \(0.300\) \\
Tiny-1.1B & \(0.327\) & \(0.370{\pm}0.004\) & \(0.358{\pm}0.004\) & \(0.300\) \\
\midrule
Macro & \(0.336\) & \(0.386\) & \(0.372\) & \(0.299\) \\
\bottomrule
\end{tabular*}
}
\caption{ARC-Challenge results for different language models. \(R_{\mathrm{score}}\) are three-seed mean$\pm$SD. Mass is normalized update mass.}
\label{tab:cross_backbone}
\end{table}

\subsection{Cross-Backbone Evaluation}
We evaluate \textsc{PARALLEL} on Qwen2.5-0.5B-Instruct, SmolLM2-360M-Instruct, and TinyLlama-1.1B-Chat using ARC-Challenge, \(\rho=0.30\), and three seeds. As shown in Table~\ref{tab:cross_backbone}, \textsc{PARALLEL} improves over Frozen across all backbones with observed update masses of \(0.297\)--\(0.300\). It retains \(95.6\)--\(96.8\%\) of Full performance and recovers \(66.1\)--\(74.5\%\) of its adaptation gain, with macro averages of \(96.2\%\) and \(71.1\%\), respectively. These results demonstrate consistent performance--update behavior across the three compact backbones.

\subsection{Ablation of Controller Inputs}

We examine the contribution of each controller input by removing the goal-related signal $s_g$, the uncertainty-related signal $s_u$, or the model representation $h_t$, while keeping the training and budget settings unchanged.

\begin{table}[t]
\centering
\setlength{\tabcolsep}{4pt}
\renewcommand{\arraystretch}{1.05}
\begin{tabular}{@{}lccc@{}}
\toprule
\textbf{Variant} & \textbf{ARC-C} & \textbf{OBQA} & \textbf{CSQA} \\
\midrule
\textsc{PARALLEL}
& $0.5691$ & $0.5400$ & $0.8419$ \\
w/o goal signal $s_g$
& $0.5529$ & $0.5340$ & $0.8288$ \\
w/o uncertainty signal $s_u$
& $0.5606$ & $0.5392$ & $0.8411$ \\
w/o representation $h_t$
& $0.5469$ & $0.5240$ & $0.8387$ \\
\bottomrule
\end{tabular}
\caption{Ablation of the controller inputs.}
\label{tab:signal_ablation}
\end{table}

As shown in Table~\ref{tab:signal_ablation}, the complete \textsc{PARALLEL} configuration achieves the highest accuracy on all three benchmarks. Removing $s_g$ causes clear degradation on ARC-Challenge and CommonsenseQA, while removing $h_t$ produces the largest reductions on ARC-Challenge and OpenBookQA. Overall, no ablated variant outperforms the complete configuration, supporting the complementary use of the controller inputs.

\begin{figure}[t]
    \centering
    \includegraphics[width=0.43\textwidth]{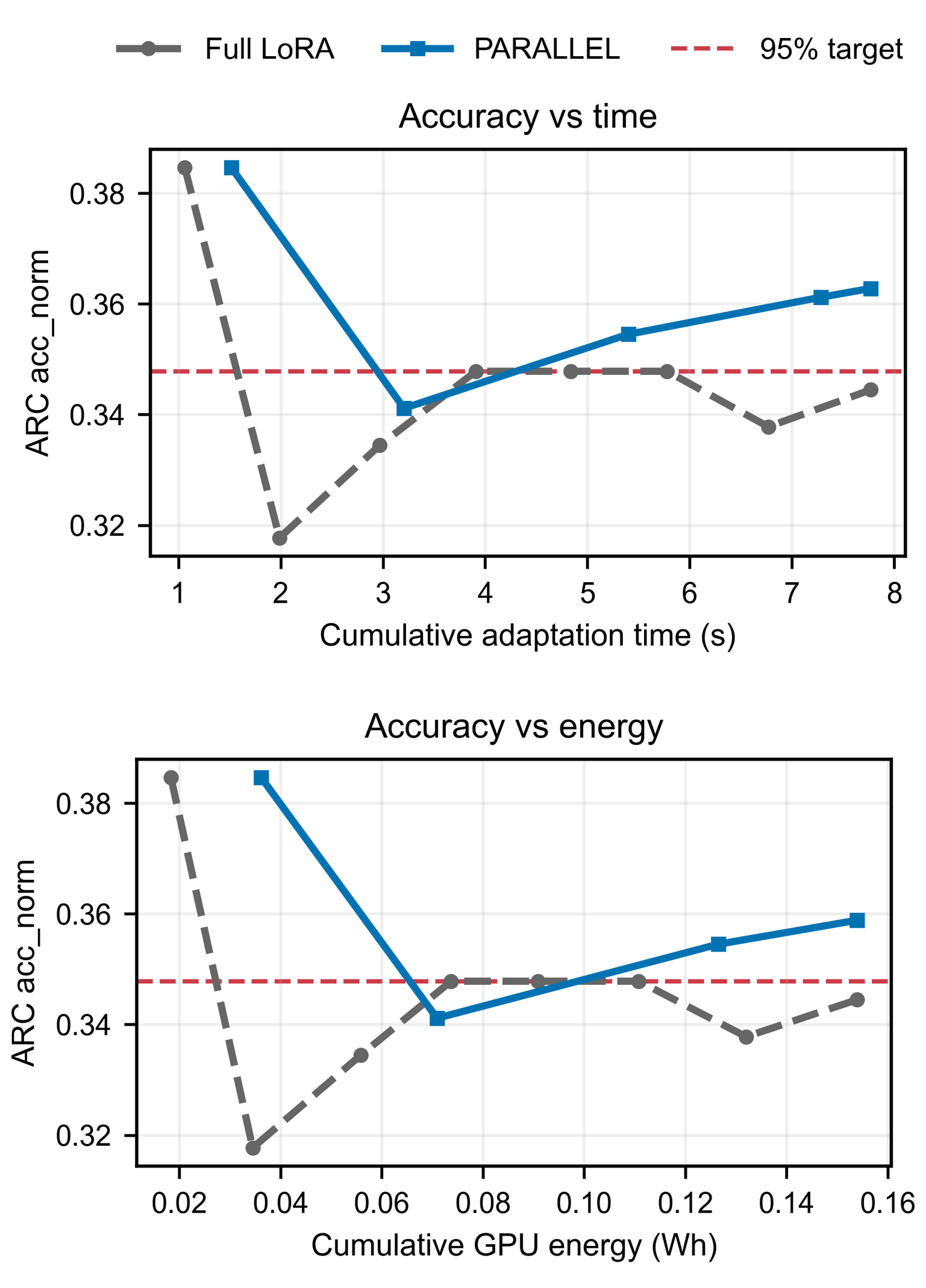}
    \caption{Validation accuracy--cost trade-offs under sequential sample updates, measured by cumulative adaptation time and GPU energy.}
    \label{fig:stream_efficiency}
\end{figure}

\subsection{Practical Efficiency of PARALLEL}
 We evaluate whether sample-wise update allocation yields higher target-task accuracy when post-deployment adaptation is constrained by time or GPU energy. Unlike the preceding endpoint comparisons, this experiment compares validation accuracy at the same measured adaptation cost.
 
 The current adapter is evaluated every 64 samples on all 299 validation examples. Adaptation time includes method-specific stream processing and updates, while GPU energy is obtained by integrating NVIDIA Management Library (NVML) power measurements sampled every 50~ms~\cite{kasichayanula2012power}. 

Figure~\ref{fig:stream_efficiency} presents a representative seed-42 trajectory on a sequential 512-sample ARC-Challenge stream. Both Full LoRA and \textsc{PARALLEL} process every incoming sample. Full LoRA updates the adapter on every sample, whereas \textsc{PARALLEL} uses the controller to determine whether and how strongly the adapter is updated. We measure \texttt{acc\_norm} and cumulative adaptation costs at fixed intervals. Cumulative cost denotes the running time or energy from the beginning of the stream and includes stream processing, controller computation, and adapter updates, while excluding checkpoint evaluation.

The top and bottom panels report accuracy according to cumulative adaptation time and GPU energy, respectively. A higher curve at the same cumulative cost indicates greater adaptation accuracy under the same resource limit. At approximately \(7.8\) seconds, \textsc{PARALLEL} achieves \(0.363\), compared with \(0.344\) for Full LoRA. At approximately \(0.155~\mathrm{Wh}\), the corresponding scores are \(0.359\) and \(0.344\). These results demonstrate a more favorable accuracy--cost balance for resource-constrained post-deployment adaptation.

\section{Conclusion}

We introduced \textsc{PARALLEL}, a prefrontal-aligned framework for sample-wise LoRA update allocation under a cumulative update-mass constraint. Its controller combines goal-related, uncertainty-related, and representation signals to determine whether and how strongly to update each labeled sample and is trained with one-step REINFORCE using immediate utility--cost feedback.
Experiments across three reasoning benchmarks and two summarization tasks show that \textsc{PARALLEL} retains most of the performance under Full adaptation while operating under a limited cumulative update mass.
It also achieves higher accuracy than Full LoRA at matched cumulative adaptation time and GPU energy in the stream setting. Controller-input ablations further support the complementary use of all three controller inputs for sample-wise update allocation.

These results demonstrate the potential of sample-wise update allocation for language-model adaptation under a limited cumulative update mass. Future work will extend \textsc{PARALLEL} to larger models, continual and generative adaptation, and broader deployment conditions while directly optimizing model stability and measured hardware costs.


\bigskip
\newpage

\bibliography{aaai2027}


\end{document}